\documentclass[letterpaper]{article} 
\usepackage{aaai24}  
\usepackage{times}  
\usepackage{helvet}  
\usepackage{courier}  
\usepackage[hyphens]{url}  
\usepackage[pagebackref,breaklinks,colorlinks]{hyperref}
\usepackage{graphicx} 
\usepackage{natbib}  
\usepackage{caption} 
\usepackage{algorithm}
\usepackage{algorithmic}
\usepackage{bbding}
\usepackage{siunitx}

\usepackage{tabularx}
\usepackage{booktabs}
\usepackage{newfloat}
\usepackage{listings}
\DeclareCaptionStyle{ruled}{labelfont=normalfont,labelsep=colon,strut=off} 
\floatstyle{ruled}
\newfloat{listing}{tb}{lst}{}
\floatname{listing}{Listing}
\title{VIGC: Visual Instruction Generation and Correction}

\author{
   Bin Wang$^{* 1}$, 
   Fan Wu$^{* 1}$, 
   Xiao Han$^{* 1}$, 
   Jiahui Peng$^{* 1}$, 
   Huaping Zhong$^{* 2}$, \\
   Pan Zhang$^{1}$, 
   Xiaoyi Dong$^{1,3}$, 
   Weijia Li$^{4}$,
   Wei Li$^{1}$, 
   Jiaqi Wang$^{1}$, 
   Conghui He$^{\dag 1}$
}
\affiliations{
    \textsuperscript{\rm 1}Shanghai AI Laboratory, \\
    \textsuperscript{\rm 2}SenseTime Research,
    \textsuperscript{\rm 3}The Chinese University of Hong Kong,        
    \textsuperscript{\rm 4}Sun Yat-sen University\\
    \{wangbin,wufan,hanxiao,pengjiahui,zhangpan,dongxiaoyi,liwei,wangjiaqi,heconghui\}@pjlab.org.cn \\
    zhonghuaping@sensetime.com, liweij29@mail.sysu.edu.cn
}

\usepackage{bibentry}
\usepackage{subfiles}
\usepackage{multirow}
\usepackage{booktabs}

\begin{document}

\maketitle
\begin{abstract}

The integration of visual encoders and large language models (LLMs) has driven recent progress in multimodal large language models (MLLMs). However, the scarcity of high-quality instruction-tuning data for vision-language tasks remains a challenge. The current leading paradigm, such as LLaVA, relies on language-only GPT-4 to generate data, which requires pre-annotated image captions and detection bounding boxes, suffering from understanding image details. A practical solution to this problem would be to utilize the available multimodal large language models to generate instruction data for vision-language tasks. However, it's worth noting that the currently accessible MLLMs are not as powerful as their LLM counterparts, as they tend to produce inadequate responses and generate false information. As a solution for addressing the current issue, this paper proposes the Visual Instruction Generation and Correction (VIGC) framework that enables multimodal large language models to generate instruction-tuning data and progressively enhance its quality on-the-fly. Specifically, Visual Instruction Generation (VIG) guides the vision-language model to generate diverse instruction-tuning data. To ensure generation quality, Visual Instruction Correction (VIC) adopts an iterative update mechanism to correct any inaccuracies in data produced by VIG, effectively reducing the risk of hallucination. Leveraging the diverse, high-quality data generated by VIGC, we finetune mainstream models and validate data quality based on various evaluations. Experimental results demonstrate that VIGC not only compensates for the shortcomings of language-only data generation methods, but also effectively enhances the benchmark performance. The models, datasets, and code are available at \url{https://opendatalab.github.io/VIGC}.

\end{abstract}

\section{Introduction}

Over the past year, there have been significant advancements in language models, particularly with the emergence of instruction tuning within Large Language Models (LLMs). This technology enables models to perform complex tasks in a zero-shot manner \cite{chatgpt, openai2023gpt4}. The fusion of visual encoders with these LLMs \cite{touvron2023llama, chiang2023vicuna} has led to substantial strides in the field of multimodal LLMs, resulting in the creation of frameworks such as BLIP-2 \cite{li2023blip}, MiniGPT-4 \cite{zhu2023minigpt}, LLaVA \cite{liu2023visual}, InstructBLIP \cite{dai2023instructblip} and InternLM-XComposer~\cite{zhang2023internlm}. These frameworks have propelled the rapid evolution of image-text multimodal tasks, exhibiting impressive capabilities in image-text dialogue.

\begin{figure}[t]
\begin{center}
	\includegraphics[width=0.98 \linewidth]{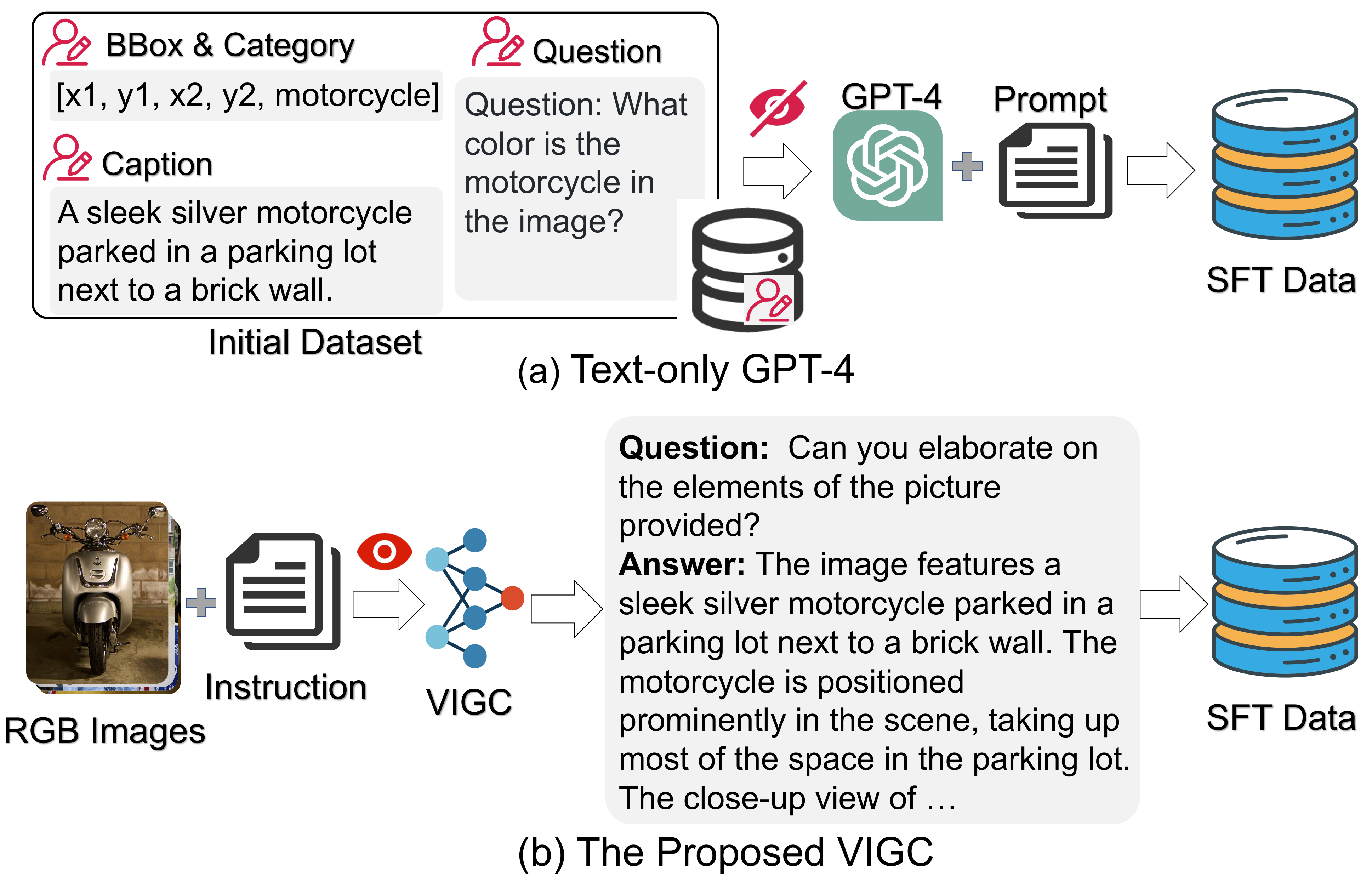}
\end{center}
		\caption{Comparison between the language-only GPT-4 approach and the proposed method, highlighting two key limitations of the former: (1) The necessity for extensive human annotation, and (2) The inability to process visual data, resulting in a loss of detailed information.}
\label{fig:fig0_instruction}
\end{figure}

Traditional multimodal models follow a two-stage training process. The initial stage involves training the model with image-text pairs to enhance feature alignment between the two modalities. The subsequent stage utilizes high-quality multimodal instruction tuning data to augment the model's ability to follow instructions, thereby improving its response to user inquiries. However, compared to a large amount of available multimodal pre-training data \cite{schuhmann2022laion, sharma2018conceptual, changpinyo2021conceptual,he2023wanjuan}, acquiring high-quality instruction tuning data is relatively more challenging. Current high-quality multimodal fine-tuning data \cite{liu2023visual, li2023mimic} is primarily generated based on language-only GPT-4 \cite{openai2023gpt4} as illustrated in Figure~\ref{fig:fig0_instruction}-(a). This approach necessitates costly manual pre-annotation and restricts the design of questions and generated responses to existing annotated information. Consequently, if the question posed is not within this annotated information, GPT-4 is unable to respond. This method also loses the detailed information in the image for answerable questions. 

To address this issue, researchers have started to consider generating data with Vision-Language Models (VLMs) \cite{zhu2023chatgpt, you2023idealgpt, zhang2023internlm} as VLMs have seen a vast amount of image-text pairs during the pre-training phase and inherently contain a wealth of knowledge. Currently, the accessible MLLMs are less powerful than their LLM counterparts. They often produce inadequate responses and generate false information, e.g., hallucination. However, existing approaches attempt to generate data using VLMs without considering how to ensure the quality of the generated data or validate it experimentally.

\begin{figure*}[ht]
\begin{center}
	\includegraphics[width=1.0 \linewidth]{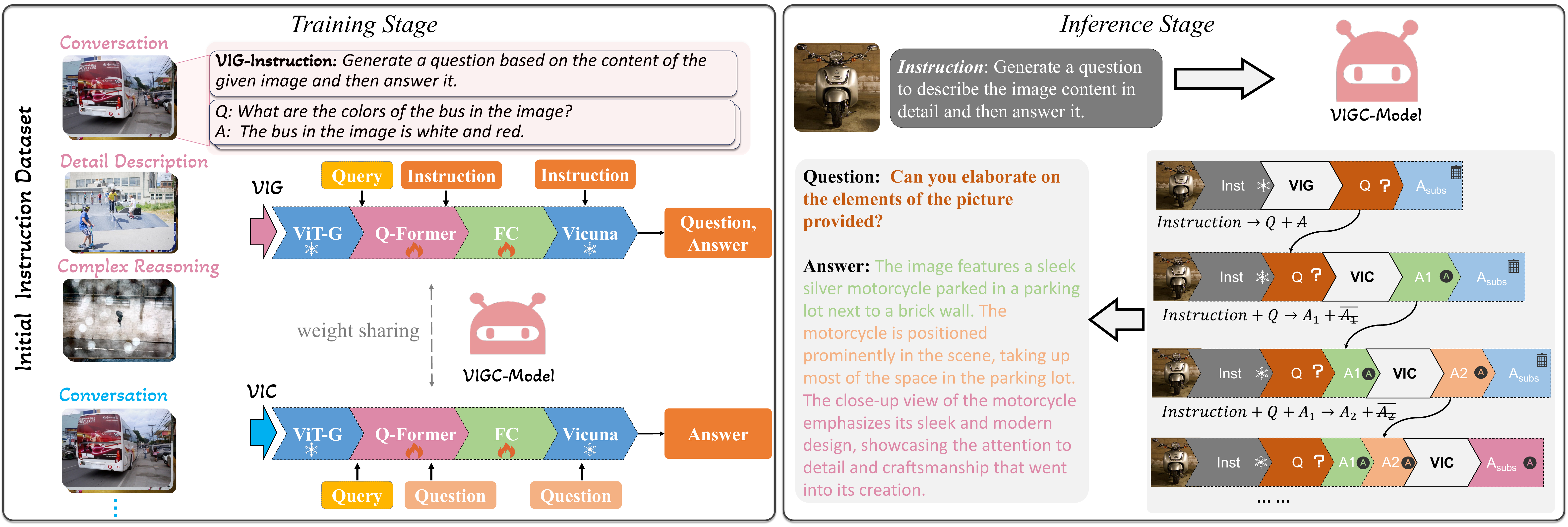}
\end{center}
		\caption{The proposed Visual Instruction Generation and Correction (VIGC) framework. The left panel illustrates the VIGC training process: Instruction fine-tuning data is fed into the VIG and VIC sub-modules. VIG aims to generate image-related question-answer pairs, while VIC refines the VIG-produced answers for precision. The right panel depicts the inference phase, where VIGC takes an arbitrary image as input, generates initial answers, and then refines them to construct high-quality data.}
\label{fig:fig1_overview}

\end{figure*}

In contrast to the aforementioned methods, we propose Visual Instruction Generation and Correction, a new method for high-quality instruction data generation. This method, based on existing visual-language models, guides the model to generate diverse visual-language question-answer pairs on new images through the fine-tuning of initial instruction data. The ability to generate diverse data is derived from the fact that both the visual encoder and the large language model have been fine-tuned on extensive datasets, encompassing rich image understanding and logical language capabilities. However, we found that data generated directly from provided instructions suffer from severe hallucination issues, which is a common problem plaguing large multimodal models \cite{peng2023kosmos,liu2023aligning,zhao2023beyond,huang2023opera}. Fortunately, our visual instruction correction module can significantly reduce model hallucination phenomena through iterative updates. The primary contributions of this study include:

\begin{itemize}
  \item We introduce Visual Instruction Generation and Correction (VIGC), a framework capable of autonomously generating high-quality image-text instruction fine-tuning datasets. The VIGC framework consists of two sub-modules: Visual Instruct Generation (VIG) and Visual Instruct Correction (VIC). Specifically, the VIG generates initial visual question-answer pairs, while VIC mitigates model hallucination and obtains high-quality data through an Iterative Q-Former (IQF) update strategy.
  
  \item We release a series of datasets\footnote{\url{https://opendatalab.com/OpenDataLab/VIGC-InstData}}~\cite{conghui2022opendatalab} generated using VIGC, including 36,781 VIGC-LLaVA-COCO and approximately 1.8 million VIGC-LLaVA-Objects365, for research on large multimodal models. To the best of our knowledge, this is the first-ever multimodal instruction fine-tuning dataset autonomously generated by a MLLM.

  \item We have conducted extensive experiments on the generated data. When trained in conjunction with the VIGC-generated data, the performance of the LLaVA-7B model significantly improved, even surpassing that of the LLaVA-13B model. Furthermore, on mainstream multimodal evaluation datasets such as MMBench, OKVQA, and A-OKVQA, models trained with the VIGC data uniformly demonstrated performance enhancements.

\end{itemize}

\section{Related Work}

\subsection{Instruction-following LLMs}
The domain of Natural Language Processing (NLP) has been significantly shaped by the advent and evolution of large language models (LLMs), including but not limited to GPT-3 \cite{brown2020language}, PaLM \cite{chowdhery2022palm}, T5 \cite{raffel2020exploring}, and OPT \cite{zhang2022opt}. These models, equipped with extensive training data and sophisticated optimization techniques, have demonstrated remarkable performance across various tasks. However, a notable challenge persists in their ability to effectively follow instructions, often leading to suboptimal results in diverse real-world applications. Efforts to address this issue have led to the introduction of various instruction fine-tuning datasets. Enhanced models, such as InstructGPT \cite{ouyang2022training}, ChatGPT \cite{chatgpt}, FLAN-T5 \cite{chung2022scaling}, FLAN-PaLM \cite{chung2022scaling}, and OPT-IML \cite{iyer2022opt}, have been developed to improve upon zero-shot and few-shot learning capabilities, primarily by learning to map instructions to the corresponding expected outputs. Despite these advancements, the generation of instruction datasets frequently relies on pre-existing NLP tasks, which curtails their generalizability. To augment the quality and diversity of instructions, Wang et al. \cite{wang2022super} introduce SELF-INSTRUCT, a methodology that employs generated instruction data to enhance the performance of LLMs. While these methods have made significant strides in augmenting the instruction-following capabilities of language models, they exhibit a standard limitation in that they cannot be directly generalized to multimodal data. 

\subsection{Multi-modal Instruction Tunning}

Compared to creating language instruction fine-tuning datasets, constructing multimodal instruction fine-tuning datasets requires a thorough understanding of image content and the development of the corresponding text. MiniGPT-4 utilizes a feature-aligned model to interpret the CC dataset \cite{sharma2018conceptual, changpinyo2021conceptual}, employs ChatGPT for initial filtering, and ultimately curates 3,500 image-text pairs for model refinement. However, this methodology encounters restrictions in terms of instruction diversity and volume.
In contrast, LLaVA proposes an innovative approach based on a language-only GPT-4 \cite{openai2023gpt4} to generate multimodal instruction data from information that includes caption descriptions and target data. While this approach generates high-quality data, it demands manual annotation of each caption description, target information, and question, which inherently limits scalability.
To extend data across a more comprehensive array of tasks, InstructBLIP pioneers an Instruction template construction methodology, converting 26 unique datasets into instruction fine-tuning data and achieving impressive results across several tasks. Concurrently, MIMIC \cite{li2023mimic} assembles larger-scale instruction fine-tuning datasets. 

Nevertheless, all these datasets require human intervention in the form of annotations, and their diversity is inherently limited by the existing data. By contrast, our study aims to propose a self-guided, model-driven instruction fine-tuning data generation method, which is capable of creating high-quality fine-tuning data suitable for any novel image.

\subsection{Visual Question Generation}
Visual Question Generation (VQG) aims to generate relevant questions based on provided images, which poses considerable challenges due to the need for diversity, naturalness, and engagement. Mostafazadeh \emph{et al.} \cite{mostafazadeh2016generating} propose the task of Visual Question Generation (VQG) and attempt to establish a foundational VQG framework, employing both retrieval-based and generative methodologies.
iQAN \cite{li2018visual} later proposed a unified, reversible network addressing both VQA and VQG tasks, enabling both answer retrieval and question generation from images. Guiding models like Guiding Visual Question Generation \cite{vedd2021guiding} have also contributed significantly to the field.

This paper proposes the Visual Instruction Generation and Correction, a model that generates image-related content, similar to VQG. Unlike the existing work, our method introduces an additional layer of complexity by developing diverse questions and providing appropriate answers based on different requirement categories. Leveraging the vast knowledge of large language models, our model's output outperforms traditional VQG tasks, which are usually limited by their training sample size.

\section{Methods}

This paper concentrates on leveraging the power of existing vision-language models to generate multimodal instruction following data autonomously. The proposed approach facilitates the creation of robust and diverse fine-tuning datasets, eliminating the requirement for intensive manual intervention. However, utilizing existing multimodal models to achieve this objective presents substantial challenges. To mitigate these,  we introduce a self-instructing framework named VIGC. Guided by existing fine-tuning data, this framework can generate higher quality and more diverse new data, as depicted in Figure \ref{fig:fig1_overview}.

\subsection{Initial Instruction Construction}

In contrast to language instructions, which can be effortlessly generated by standalone language models \cite{peng2023instruction, wang2022super}, the construction of visual-text multimodal instructions requires a detailed understanding of visual content, as well as the ability to pose relevant questions and provide correct answers based on the actual content of the images. Nevertheless, existing multimodal models are deficient in their capacity to directly generate visual-language instruction data. To overcome this limitation, we exploit readily available instruction fine-tuning data and formulate additional instruction templates, thereby facilitating the automatic generation of instruction data.

\begin{figure}[ht]
\begin{center}
	\includegraphics[width=0.98 \linewidth]{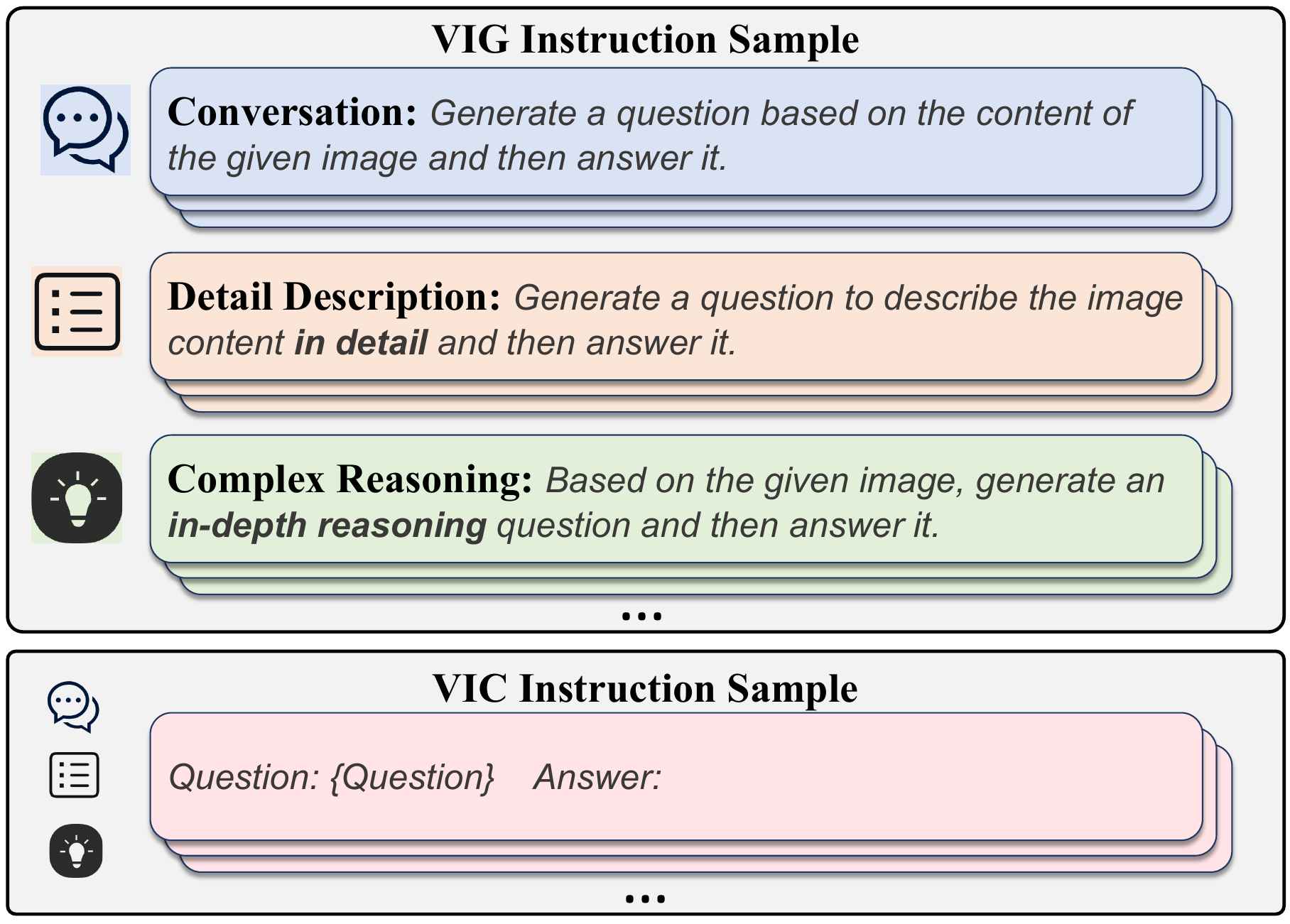}
\end{center}
		\caption{Template examples corresponding to instruction tuning in VIG and VIC submodules.}
\label{fig:fig2_instruction}
\end{figure}

Our proposed method is universally applicable to generating various types of image-text multimodal instruction fine-tuning data. To elucidate our approach, we exemplify it using the generation of LLaVA-style data instructions. Specifically, we construct instruction templates encompassing dialogue, detailed description, and complex reasoning, following the categorization of instruction fine-tuning data types as delineated in LLaVA. Figure \ref{fig:fig2_instruction} presents instances of these three types of instruction templates, which are essentially uncomplicated, principally requesting, \textit{``generate T-type question-answer pairs predicated on the image content.''} Theoretically, if a model can comply with these instruction descriptions following training, it should be proficient in generating question-answer pairs.

With the instruction templates and existing visual instruction-tuning data (i.e., Question-Answer pairs in LLaVA), we construct a comprehensive VIG instruction-tuning dataset as follows:

\begin{equation}
T_{VIG} = {(X_i, I_t, Q_i^t, A_i^t)}^{N_t}
\end{equation}

where $i \in \{1, 2, ..., N_t\}$, $N_t$ denotes the instruction type, such conversation, detailed description, etc.   $X_i$ represents an RGB image, $I_i$ represents an instruction corresponding to a specific type $t$, $Q_i^t$ is a question related to the image $X_i$ under the context of instruction $I_t$, and $A_i^t$ is the answer to the question $Q_i^t$. Our objective is to leverage this dataset for the training of models that, given a specific instruction $I_t$, can generate corresponding question-answer pairs for a given image, following the designated instruction type. Figure \ref{fig:fig1_overview} provides illustrations of the initial instruction dataset. 

Distinguished from the VIG, the VIC instruction employs an image and a query as input for its fine-tuning process, with the objective of generating precise responses. The dataset for the VIC instruction is presented below:

\begin{equation}
T_{VIC} = {(X_i, Q_i^t, A_i^t)}^{N_t}     
\end{equation}

\subsection{Visual Instruction Generation}

In alignment with current popular multimodal models such as MiniGPT-4 \cite{zhu2023minigpt} and InstructBLIP \cite{dai2023instructblip}, the architecture of the proposed VIGC can be dissected into four primary components: the visual encoder (ViT) \cite{fang2023eva}, the large language model (Vicuna) \cite{chiang2023vicuna}, the Q-Former \cite{li2023blip} for visual feature extraction, and the Fully-Connected (FC) projection for reconciling visual-language features. Functionally, the model can be further segmented into two distinctive sub-modules: the Visual Instruction Generation (VIG) module and the Visual Instruction Correction (VIC) module. It is imperative to underscore that these two sub-modules share network parameters, the primary differentiator being the data type employed for training.

The principal objective of the VIG module is to autonomously produce relevant visual question-answer pairs that correspond to a specific instructional command for any given image. 
Figure \ref{fig:fig1_overview} illustrates the process that the VIG module follows in the training phase. In the training phase, the VIG module stochastically selects an image, which is subsequently processed via a visual encoder. It generates a set of fixed visual feature embeddings. The Q-Former module, purposefully designed to be aware of instructional information, further refines these visual features. At this stage, the model employs learnable visual queries that perform self-attention operations in conjunction with the instruction. This operation is followed by a cross-attention phase with visual embeddings. This mechanism impels the visual features to concentrate on the instructional information, thereby augmenting their relevance and precision within the context of the assigned task. Following the cross-attention phase, the refined features are channeled through an FC mapping layer, a crucial step that aligns visual features with their linguistic counterparts, thereby ensuring a seamless integration of visual and language features. Subsequently, the instruction-aligned features are ingested by the language model. This process guides the model to generate the predicted results. Specifically, the objective in this context is to generate visual questions and answers that are intrinsically linked to the content of image $X_i$, the nature of which is determined by the instruction. We utilize the original auto-regressive loss function inherent to the large language model. This methodology guides the model in generating sentences that align with the question-answer pairs provided in the training set.

\subsection{Visual Instruction Correction}

In the exploration conducted for this study, we discovered that existing multimodal models \cite{liu2023visual}, \cite{dai2023instructblip}, much like language models \cite{radford2018improving, radford2019language, brown2020language, openai2023gpt4, chatgpt}, often exhibit hallucination issues. This hallucination phenomenon is also present in the data generated by the VIG, especially in instances of extensive descriptions. We attribute this to the tendency of multimodal models to progressively rely on the current answer text during the answer generation phase, thereby gradually neglecting the image information and consequently leading to the description of targets not present in the image. To eliminate the hallucination phenomenon in generated data and ensure that downstream tasks based on this data are not contaminated, we specifically introduce an instruction correction module to update the answers and reduce the occurrence of hallucinations.

To effectively utilize the VIC, specific actions need to be undertaken during both the model training and inference stages:

During the training phase: The goal of the VIG phase is to generate corresponding visual question-answer pairs given an instruction. Conversely, the objective of the VIC training phase is to supply the model with a Question, thereby directing the model to focus on extracting features pertinent to the input question/text during the Q-Former feature extraction process. These features lay the groundwork for subsequent answers.

During the inference phase: After training the model using the aforementioned VIC method, it can take the questions from the question-answer pairs generated by the VIG as input and regenerate answers. Since the model places greater emphasis on the question when formulating responses, the generated results are typically more accurate. Furthermore, we iterate this Q-Former feature updating process, termed as Iterativate-Q-Former (IQF), as illustrated in the VIGC inference phase in Figure~\ref{fig:fig1_overview}. Before deploying the VIC module, we initially generate the initial question (Q) and answer (A) using the VIG. In the first iteration, we use the Instruction and Question as inputs to output answers $A_1$ and $\bar{A}_1$, where $A_1$ represents the first sentence of the answer and $\bar{A}_1$ signifies all content following the first sentence. In the second iteration, we input the Instruction, Question, and the answer $A_1$ from the previous step to predict $A_2$, and this process continues iteratively until a termination symbol is encountered. The efficacy of this iterative approach is primarily due to the continual updating of visual features with the most recent textual information, making subsequent results more accurate. However, it should be noted that while this method is highly beneficial for providing detailed descriptions of image content, its effectiveness for dialogue tasks and inference tasks is relatively limited. This is because dialogue tasks usually consist of single sentences, and the subsequent content in inference tasks does not heavily depend on image information.

\begin{figure*}[!ht]
\begin{center}
        \includegraphics[width=0.95 \linewidth]{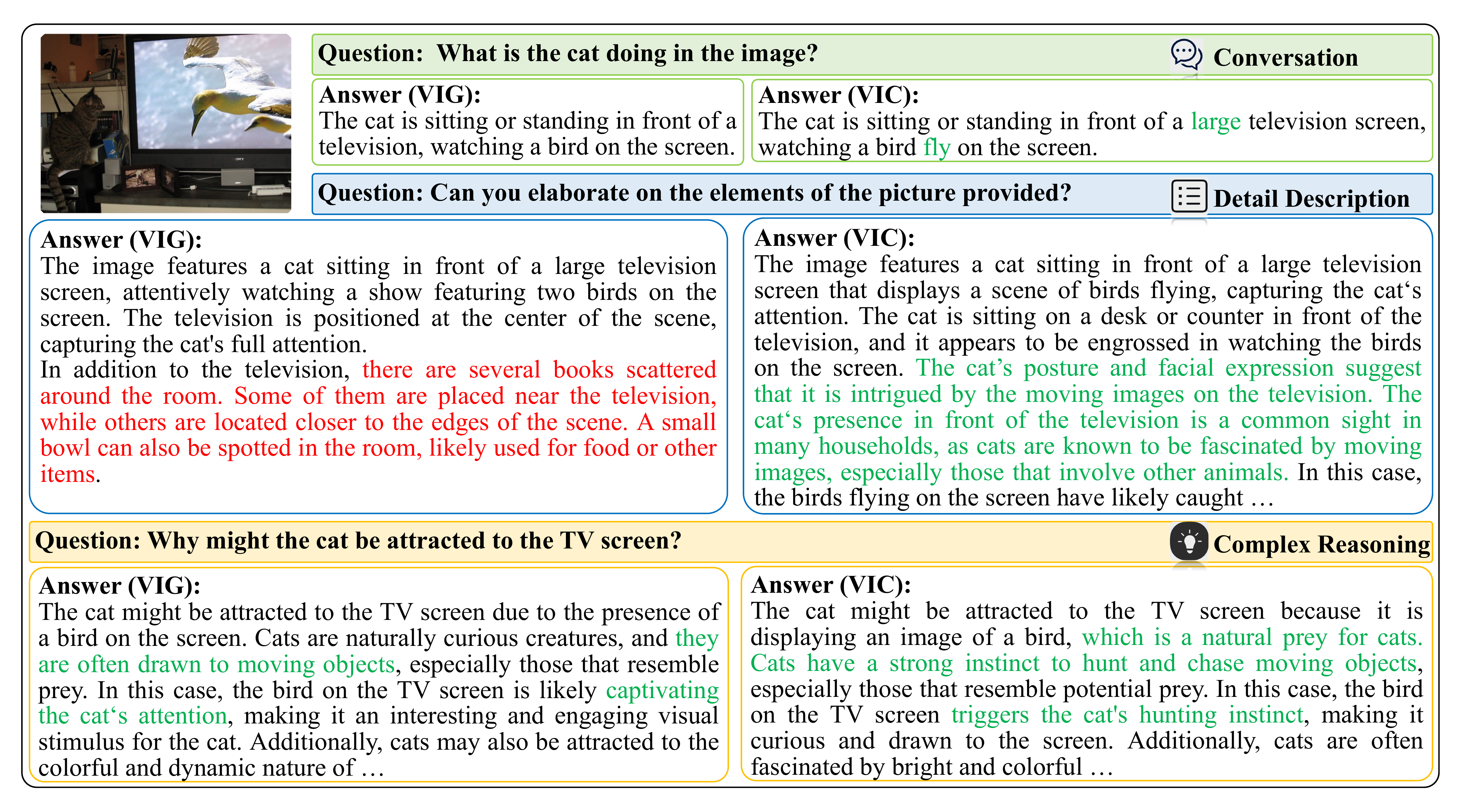}

\end{center}
  \caption{Generated Instructions based on the proposed VIGC.}
\label{fig:fig4_vigc_data}
\end{figure*}

\section{Experiments}

\subsection{Datasets}

\textbf{Training Data}. We trained the VIGC network using two types of visual-language instruction fine-tuning data. The first type, represented by the LLaVA dataset \cite{liu2023visual}, is manually curated and combined with language-only GPT-4 \cite{openai2023gpt4} for multimodal models. It includes 150K training samples, subdivided into simple dialogue (57,669 samples), detailed description (23,240 samples), and complex reasoning vision-language data (76,803 samples). This dataset spans various facets of multimodal dialogue, including category recognition, counting, action recognition, and scene recognition. The detailed descriptions demand careful image observation and comprehensive detailing, while the complex reasoning tasks require deep inference and external knowledge integration.
The second type of data is multimodal instruction fine-tuning data derived from publicly available image-text datasets. Specifically, we used OKVQA \cite{marino2019ok} and A-OKVQA \cite{schwenk2022okvqa} datasets, as utilized in InstructBLIP \cite{dai2023instructblip}, for VIGC training. These datasets, necessitating extensive external knowledge, are ideal for assessing the VIGC's capabilities.

\noindent\textbf{Inference Data}. Following the VIGC network training, we generated fine-tuning data for multimodal instruction using image datasets. We employed two distinct datasets, COCO \cite{lin2014microsoft} and Objects365 \cite{shao2019objects365}, to evaluate VIGC's effectiveness in handling data within the same or different image domains. The COCO dataset serves as the foundation for the construction of the LLaVA, OKVQA, and A-OKVQA datasets. It's crucial to emphasize that during the data generation phase, we intentionally omitted any images that were previously included in the test set to ensure the fairness and effectiveness of the evaluation.

\subsection{Implementation Details}

During the training phase of VIGC, we utilize the MiniGPT-4 \cite{zhu2023minigpt} first-stage pre-trained model as the source of initial parameters. This ensures that the initial model does not incorporate additional instruction fine-tuning data for training, thereby preserving the fairness of downstream task validation. This model encompasses the ViT-G/14 from EVA-CLIP \cite{fang2023eva}, the Q-Former \cite{li2023blip}, and a linear projection layer. The language models employed are Vicuna7B and Vicuna13B  \cite{chiang2023vicuna}. It is noteworthy that, as illustrated in Figure 1, our Q-Former is designed to receive either Instruction or Question text simultaneously, which is crucial for the iterative correction in VIC. Therefore, we utilize the Q-Former from BLIP2-FlanT5\textsubscript{XXL}~\cite{li2023blip} as the initial parameters for the Q-Former. We designate this network model as MiniGPT-4+. During the training process, only the parameters of the Q-Former and the linear projection layer are subjected to fine-tuning, while the parameters of the language and visual models remain constant. The training is conducted throughout $10$ epochs, with the model's performance being validated after each epoch. The model that demonstrates the best performance is subsequently selected for data generation.

In terms of batch sizes, we utilize 64 for both 7B and 13B models. The entire training process, executed on 8 A100 (80GB) GPUs, completes in approximately 10 hours. 

\subsection{LLaVA Data and Evaluation}

\textbf{Dataset Analysis}. In the pursuit of generating a more diverse set of LLaVA-like data, the VIGC model is trained using a combination of LLaVA-150K data and three types of instruction templates. During the inference phase, we utilized images from the COCO 2017 training set, intentionally excluding those already included in the LLaVA dataset. This resulted in the selection of a total of 36,781 initial images, which served as the foundation for instruction data generation; we refer to this data as \textbf{coco-extra}, which serves as the default supplementary data used for model training during evaluation.

Based on the aforementioned data, the VIG network generates diverse initial questions and answers. Subsequently, the VIC network refines the outputs by taking the questions and the existing answers as inputs through the Iterative Q-Former (IQF) operation, thus generating higher-quality responses. Figure~\ref{fig:fig4_vigc_data} illustrates the three categories of data generated via the VIGC process:

\begin{itemize}

    \item Conversation: The questions are typically specific, eliciting concise and clear responses. 

    \item Detail Description: The questions are relatively fixed and revolve around describing the image's content. This requires the model to clearly observe all targets within the image. It is observed that the detailed descriptions generated directly from Visual Instruction Generation (VIG) are fraught with numerous illusions. However, after the application of Visual Instruction Correction (VIC), these illusory phenomena have significantly diminished.

    \item Complex Reasoning: The posed questions necessitate the integration of external knowledge and the application of sophisticated logical reasoning skills. 
\end{itemize}

\begin{table}[t]
    \small
	\centering
    \resizebox{0.48\textwidth}{!}{%
	\begin{tabular}{lcccc} 
	\toprule
    Model & \multicolumn{4}{c}{Metrics} \\
    \cmidrule(lr){2-5}
    & Conv & Detail & Complex & All \\
	\midrule
    LLaVA-7B (Baseline)       & 75.1 & 75.4 & 92.3 & 81.0 \\
	\midrule
    add 36K Conv (VIG)            & 80.9 & 76.1 & 92.6 & 83.3 \\
    add 36K Conv (VIC)            & 83.9 & 76.9 & 90.9 & \textbf{84.0} \\
	\midrule
    add 36K Detail (VIG)          & 80.2 & 72.7 & 90.9 & 81.4 \\
    add 36K Detail (VIC)          & 83.3 & 80.6 & 93.1 & \textbf{85.8}  \\
	\midrule
    add 36K Complex (VIG)         & 81.4 & 75.6 & 90.5 & 82.6 \\
    add 36K Complex (VIC)         & 80.2 & 76.2 & 93.2 & \textbf{83.3} \\
    \midrule
    replace 10K Conv             & 78.2 & 76.5 & 91.6 & 82.1 \\
    replace 10K Detail           & 75.8 & 79.8 & 91.2 & 82.2 \\
    replace 10K Complex          & 77.5 & 77.8 & 92.8   & 82.8 \\
    replace Conv/Detail/Complex  & 78.3 & 76.6 & 92.4 & 82.5 \\ 
	\bottomrule
	\end{tabular}
    }
	\caption{Comparative Evaluation of VIGC Data Addition vs. Replacement in Model Training on the LLaVA Evaluation}	
	\label{tab:vigc-coco1}
\end{table}

Overall, the quality of the visual-language question-answer pairs autonomously generated by the model has exceeded our initial expectations. We posit that this rich new knowledge inherently resides within the language model itself, and we have merely employed multimodal instruction fine-tuning to distill this knowledge onto new multimodal data.

\begin{table*}[ht!]
    \centering
    \resizebox{0.95\textwidth}{!}{%
    \begin{tabular}{lccccccc|cccc}
    \toprule
    & \multicolumn{7}{c|}{MMBench} & \multicolumn{4}{c}{LLaVA} \\
    Method  & LR & AR & RR & FP-S & FP-C & CP & Overall & Conv & Detail & Complex & All \\
    \midrule
    MiniGPT-4+        & 10.0 & 31.3 & 7.83 & 18.9 & 13.1 & 43.0 & 24.4 & 83.5 & 77.8 & 92.4 & 84.7 \\
    MiniGPT-4+ w/ coco   & 11.7 & 27.8 & 19.1 & 27.9 & 11.0 & 44.3 & \textbf{27.5}($\uparrow$ 3.1) & 84.1 & 84.1 & 92.7 & \textbf{87.0}($\uparrow$ 2.3) \\
    \bottomrule
    \end{tabular}
    }
    \caption{Performance of MiniGPT-4+ Models on MMBench and LLaVA-Eval Datasets. MMBench Metric include Logic Reasoning (LR), Attribute Reasoning (AR), Relation Reasoning (RR), Fine-grained Perception at Instance-level (FP-S), Fine-grained Perception at Cross-instance (FP-C), and Coarse Perception (CP).}
    \label{tab:minigpt4}
\end{table*}

\begin{table}[ht]
    \centering
    \small
    \resizebox{0.48\textwidth}{!}{%
    \fontsize{10pt}{12pt}\selectfont
    \begin{tabular}{
        l
        S[table-format=2.1]
        S[table-format=2.1]
        S[table-format=2.1]
        S[table-format=2.1]
    }
    \toprule
    {Model} & {Conv} & {Detail} & {Complex} & {All} \\
  \midrule
    LLaVA-7B & 75.1 & 75.4 & 92.3 & 81.0 \\
    LLaVA-7B w/ coco & 83.3 & 80.6 & 93.1 & \textbf{85.8}{($\uparrow$ 4.8)} \\
    LLaVA-7B w/ objects365 & 86.8 & 77.6 & 90.9 & 85.2 \\
    \midrule
    LLaVA-13B* & 82.7 & 76.6 & 94.8 & 84.8 \\
    LLaVA-13B w/ coco & 88.9 & 77.4 & 93.5 & \textbf{86.8}{($\uparrow$ 2.0)} \\

    \bottomrule
    \end{tabular}
    }
    \caption{Relative scores for different settings w.r.t. GPT-4 (language-only) on LLaVA-Eval Dataset. The results for LLaVA-13B are reproduced from \cite{liu2023visual}.
    }
    \label{tab:coco_13B}
\end{table}

\begin{table}[ht]
    \small
	\centering
   \resizebox{0.45\textwidth}{!}{%
    	\begin{tabular}{lcc} 
    	\toprule
          Model & OKVQA & A-OKVQA \\
    	\midrule
        PaLM-E \cite{driess2023palm}  & 66.1 & -  \\
        PromptCap \cite{hu2022promptcap}    & 60.4 & 56.3  \\
        MiniGPT-4+ w/o VIGC    & 59.1 & 58.3  \\
        MiniGPT-4+  w/ VIGC     & \textbf{59.8}{($\uparrow$ 0.7)} & \textbf{58.9}{($\uparrow$ 0.6)} \\
        InstructBLIP  w/o VIGC               & 63.1 & 62.5  \\
        InstructBLIP w/ VIGC      & \textbf{63.8} {($\uparrow$ 0.7)} &\textbf{64.1} {($\uparrow$ 1.6)}   \\ 
    	\bottomrule
    	\end{tabular}
   }
	\caption{Results of finetuning MiniGPT-4+ and InstructBLIP on OKVQA and A-OKVQA dataset.}
	\label{tab:vqa}
\end{table}

\noindent\textbf{Dataset Evaluation}. Based on the generated data, we conducted detailed ablation experiments on LLaVA-7B to verify the performance improvement of the model after training with the generated data. The evaluation method used here is the quantitative evaluation proposed by LLaVA, where GPT-4 assesses the quality of the model's responses to given evaluation questions, which can be understood as relative scores compared to GPT-4. LLaVA provides 30 test images, each containing three types of questions, for 90 questions.

Table \ref{tab:vigc-coco1} presents the results of augmenting the original LLaVA-150K dataset with three types of generated data, followed by fine-tuning the LLaVA first-stage model with instructions. Including instruction data directly generated from VIG during the training phase has proven to be beneficial. We observed a marginal improvement when adding detailed description data generated by VIG, which can be attributed to the severe illusions present in this data. In contrast, the incorporation of conversation data and complex reasoning data has led to appreciable performance gains.

Further refining the data using VIC and then training the model with the augmented conversation data, detailed description data, and complex reasoning data resulted in additional improvements. The performance metrics have reached $84.0\%$, $85.8\%$, and $83.3\%$, respectively. These results underscore the critical role of VIC in eliminating hallucinations, thereby enhancing the model's overall performance. Simultaneously, to validate the superiority of the VIGC-generated dataset over the LLaVA dataset, we conducted an experiment where we randomly replaced 10,000 instances from each type of data, as well as a complete replacement of all three types of data. The experimental results indicated that, under the condition of constant data volume, the performance of the model trained on a mixture of the LLaVA dataset and the VIGC dataset surpasses that of the model trained solely on the LLaVA dataset. 

Table \ref{tab:coco_13B} presents experiments conducted on different datasets and models of varying sizes, substantiating that the use of generated data from different domains, such as Objects365 and COCO, can still lead to remarkable performance improvements. This offers a novel solution for enhancing the performance of cross-domain tasks. We also conducted experiments on LLaVA-13B, proving that performance enhancement can be achieved on larger models as well.

We also evaluated the performance of the VIGC model on MMBench, LLaVA (as shown in Table ~\ref{tab:minigpt4}) and further fine-tuned the VIGC model based on 36K COCO data generated by VIGC. We discovered that following this self-iterative training process, the model performance improved on both MMBench and LLaVA. This promising capability of self-enhancement through iterative training is a subject we plan to continue exploring in our future research.

\subsection{OK-VQA Dataset and Evaluation}

To further assess the quality of the data generated by the VIGC model, we conducted training and evaluation on the OKVQA dataset, which requires external knowledge. Specifically, we trained the VIGC network using the OKVQA dataset and corresponding instruction templates. Subsequently, we generated additional instruction fine-tuning data based on VIGC on COCO. Ultimately, we fine-tuned InstructBLIP based on OKVQA and the generated data. We found that despite InstructBLIP already utilizing a substantial amount of data in the instruction fine-tuning phase, the use of additional generated data for downstream task fine-tuning still enhanced the model's performance on specific datasets. We performed the same experimental validation on A-OKVQA.

The experimental results are presented in Table \ref{tab:vqa}. It can be seen that the performance of the InstructBLIP model, when fine-tuned with the addition of generated data, outperforms the model only fine-tuned with original data. There were improvements of 0.7\% and 1.6\% on OKVQA and A-OKVQA, respectively, achieving state-of-the-art results for models of this scale on both datasets. BUsing the MiniGPT-4+ pre-training model, we arrived at similar conclusions. This demonstrates that generated data can effectively enhance downstream fine-tuning performance, a finding that holds significant value for domains where data acquisition is challenging.


\section{Conclusion}

In this paper, we introduced the Visual Instruction Generation and Correction framework, a novel self-instruct method for autonomously generating high-quality vision-language instruction data. Leveraging the VIGC framework, we have generated diverse multimodal instruction-following data on the COCO and Objects365 datasets. The quality of this data has been validated through various evaluations. The VIGC-based approach provides a convenient means to acquire more high-quality instruction tuning data.
While using the Visual Instruct Correction has significantly reduced model hallucination, some instances persist. We intend to delve deeper into the exploration of solutions aimed at eliminating multimodal hallucinations in the future. Moreover, we are considering the potential of forming a closed-loop system by integrating the VIGC's autonomous data generation with multimodal model training. This system would enhance model performance through data improvement and, reciprocally, elevate data quality through model enhancement. 

\section{Acknowledgments}
This project was supported by National Key R\&D Program of China (No. 2022ZD0160101) and Shanghai Artificial Intelligence Laboratory.

{
\bibliography{vigc}
}

\appendix

\onecolumn 

\section*{Supplementary Material}

\subsection{A. Instruction Templates}

The instruction templates for VIGC, including conversation, detailed description, complex reasoning, and OKVQA types, are presented in Tables \ref{tab:ins_conv}, \ref{tab:ins_detail}, \ref{tab:ins_complex}, and \ref{tab:ins_okvqa} respectively.


\begin{table}[ht]
\centering
\begin{tabular}{c  p{14cm}}
\toprule
\textbf{No.} & \textbf{Instruction} \\
\midrule
1 & Generate a question based on the content of the given image and then answer it. \\
2 & Given the image, generate a question along with the answer. \\
3 & From the image provided, craft a question and answer it. \\
4 & Come up with a question related to the content of the image and provide the answer. \\
5 & Brainstorm a query associated to the image and provide the response. \\
6 & Construct a question based on the information presented in the image and answer it. \\
7 & Ask yourself a question about the content of the image and respond to it. \\
8 & Establish a query related to the content of the image and give the answer. \\
9 & Ask a question derived from the image and then answer it. \\
10 & Create a question about the image and answer it. \\
\bottomrule
\end{tabular}
\caption{The list of instructions for conversation.}
\label{tab:ins_conv}
\end{table}

\begin{table}[ht]
\centering
\begin{tabular}{c p{14cm}}
\toprule
\textbf{No.} & \textbf{Instruction} \\
\midrule
1 & Generate a question to describe the image content in detail and then answer it. \\
2 & Considering the picture, come up with a question to describe the image content in detail along with the answer. \\
3 & Describe the image content with a question and give the response. \\
4 & Come up with a creative question to express the image content and then provide the answer. \\
5 & Draft a query to address the image content and give the reply. \\
6 & Create a question to reveal the image content and give the resolution. \\
7 & Given the photo, state a question that reveals the details of the image and then answer it. \\
8 & Ask a question about what is depicted in the image and then answer it. \\
9 & Make up a query to explain the photo in more detail and answer it. \\
10 & Compose a question describing the subject of the image, followed by the answer. \\
\bottomrule
\end{tabular}
\caption{The list of instructions for detailed description.}
\label{tab:ins_detail}
\end{table}

\begin{table}[H]
\centering
\begin{tabular}{c  p{14cm}}
\toprule
\textbf{No.} & \textbf{Instruction} \\
\midrule
1 & Based on the given image, generate an in-depth reasoning question and then answer it. \\
2 & Given the image, generate an in-depth reasoning question and answer. \\
3 & Taking the image into account, generate an reasoning question along with the answer. \\
4 & Can you come up with a reasoning question based on the image and then provide the answer? \\
5 & After looking at the image, devise a reasoning question and provide the answer to it. \\
6 & Contemplate the image and create a reasoning question with the answer provided. \\
7 & Analyze the image and provide a reasoning question as well as the answer. \\
8 & Compose a reasoning question using the image with its answer. \\
9 & Evaluate the image and create a comprehensive reasoning question and its answer. \\
10 & Analyze the image and craft an effective reasoning question and its response. \\
\bottomrule
\end{tabular}
\caption{The list of instructions for complex reasoning.}
\label{tab:ins_complex}
\end{table}

\begin{table}[H]
\centering
\begin{tabular}{c  p{14cm}}
\toprule
\textbf{No.} & \textbf{Instruction} \\
\midrule
1 & Based on the content of the given image, generate a question that requires common sense to answer and then briefly answer it. \\
2 & Construct a question that draws upon common sense to answer, using the content presented in the given image, and then briefly answer it. \\
3 & Explain the content of the image in a question and then provide a short answer using knowledge types such as commonsense and facts. \\
4 & Generate a query that requires reasoning on the information depicted in the image, utilizing a variety of knowledge types like commonsense, and then offer a concise answer. \\
5 & Develop a query to demonstrate the knowledge types such as commonsense and facts related to the given image and then provide a brief answer. \\
6 & Based on knowledge types such as commonsense and facts, come up with a query related to the given image and then briefly answer it. \\
7 & Come up with a question related to the content shown in the image that requires reasoning using a variety of knowledge types such as commonsense and then succinctly answer it. \\
8 & Brainstorm a question about the content of the given image that requires reasoning with a variety of knowledge types such as common sense and then state the answer briefly. \\
9 & Construct a query that requires logic based on the contents of the given image and involves a variety of knowledge types such as commonsense, and then deliver a brief response. \\
10 & Invent an inquiry derived from the pictured material that calls for the use of different knowledge types like commonsense and subsequently summarize the solution with brevity. \\
\bottomrule
\end{tabular}
\caption{The list of instructions for outside knowledge-based visual question answering.}
\label{tab:ins_okvqa}
\end{table}


	

\begin{figure}[ht]
\begin{center}
	\includegraphics[width=0.98 \linewidth]{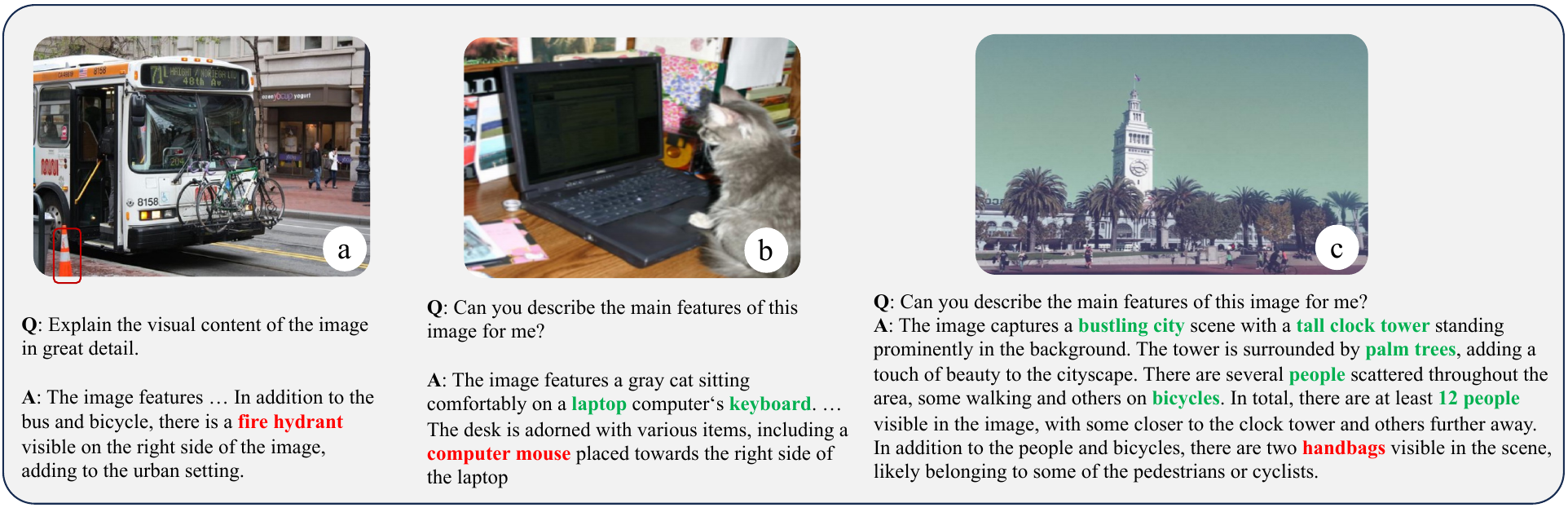}
\end{center}
		\caption{Visualization of hallucination phenomena in VIGC failure cases: (a) model limitations, (b) training data bias, and (c) Information decay in sequence generation.}
\label{fig:demo}
\end{figure}

\begin{table}[ht]
\centering
\begin{tabular}{lcccc}
\hline
Model &H. Count &  1st 50\%  & 2nd 50\% & H. Word\\
\hline
VIG   &66      & 0   & 66   & 93\\
VIC   &10      & 1   & 9   & 12 \\
\hline
\end{tabular}
\caption[Hallucination statistics for VIG and VIC models.]{Hallucination statistics for VIG and VIC models. H. Count: Hallucination Count, 1st 50\%: Hallucinations in the first 50\% of sentences, 2nd 50\%: Hallucinations in the second 50\% of sentences, H. Words: Hallucinated Words}
\label{tab:hallucination}
\end{table}

\subsection{B. Hallucination analysis for the VIGC model.}

 The ``hallucination" phenomenon in multimodal models refers to the creation of non-existent objects or details in the output. We have identified three primary causes:

\begin{itemize}
    \item \textbf{Model limitations}: Even the most advanced visual models today cannot capture all details in images and are prone to misidentifications 
    (see Fig.~\ref{fig:demo}-a).

    \item \textbf{Training data bias}: During model training, certain types of objects or scenes are more prevalent than others. Additionally, there is co-occurrence in the data, meaning that certain targets are more likely to appear in the same description/response  (see Fig.~\ref{fig:demo}-b).

    \item \textbf{Information decay in sequence generation}: In detailed description tasks, the model's output may neglect the initial image information over time, relying more on self-generated language context and creating "hallucinations"(see Fig.~\ref{fig:demo}-c and Tab.~\ref{tab:hallucination}-2nd 50\%).
    
\end{itemize}

Our VIG sub-module is designed to function as a self-questioning and answering model, distinguishing it from models that merely answer questions. It addresses the hallucination issue, particularly the third point, through iterative updates of the input image and text. This strategy helps prevent information decay and the fitting of common language patterns. As evidenced in Table~\ref{tab:hallucination}, tests conducted on 100 images from the evaluation set and synthetic data demonstrate a significant reduction in hallucination following VIC correction, decreasing from 66\% in VIG to 10\% in VIC.

\subsection{C. Data Statistical Analysis}

Based on VIGC, we generate fine-tuning data for dialogue, detailed description, and complex reasoning tasks using both COCO and Objects365 images. Each image and its corresponding question-answer pair are treated as an instance. The sample data, the average length of questions and answers, and the diversity of questions are presented in Table \ref{tab:q_diversity}.

\begin{figure}[H]
\begin{center}
	\includegraphics[width=0.72 \linewidth]{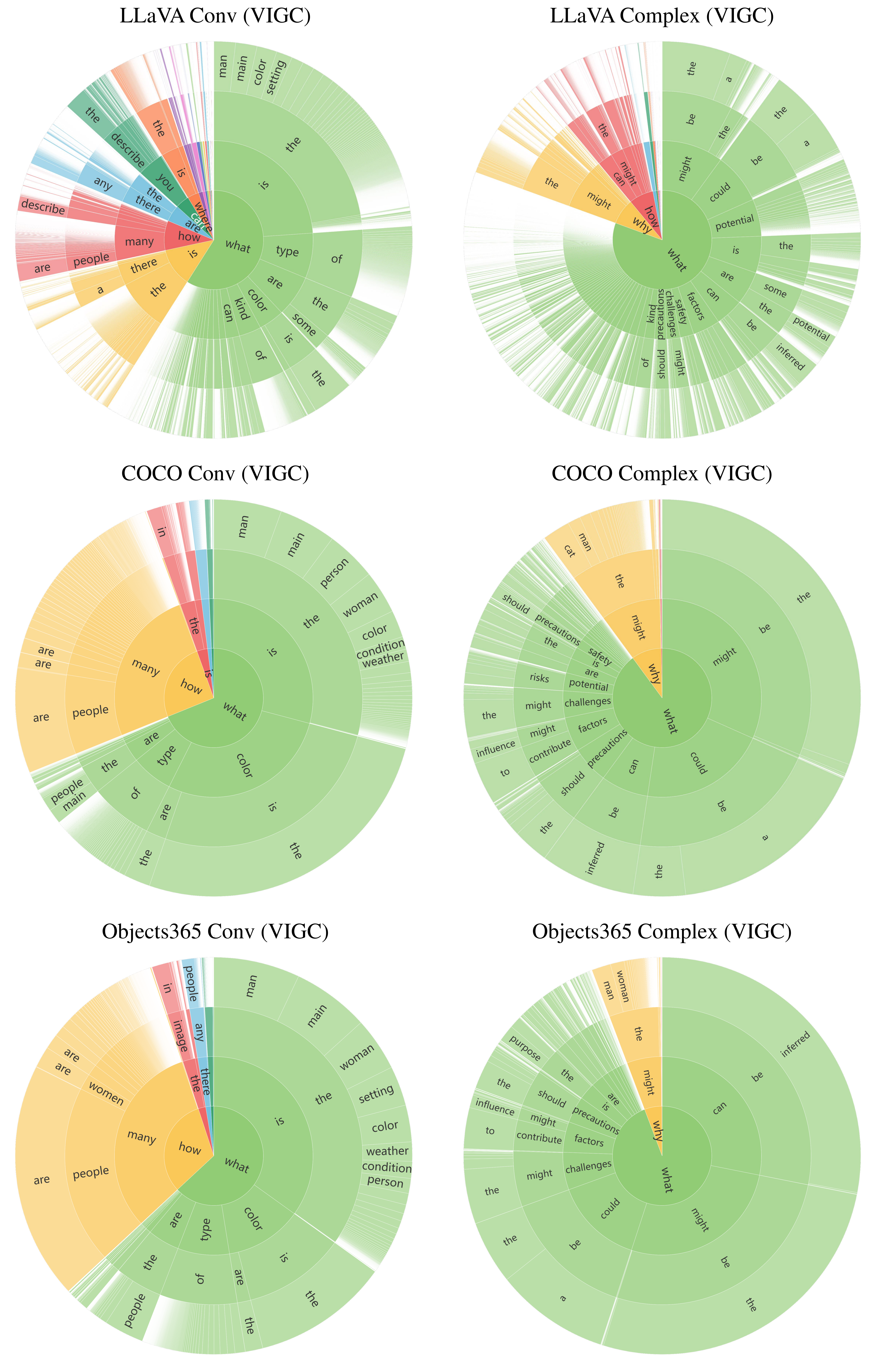}
\end{center}
		\caption{Distribution of question words in generated conversation and complex reasoning data on LLaVA, COCO, and Objects365.}
\label{fig:question_distribution}
\end{figure}

\textbf{Question Diversity.} Figure \ref{fig:question_distribution} visualizes the frequency of questions beginning with different words, illustrating the diversity of questions. Many dialogue questions begin with phrases like "what is" or "what color", whereas complex reasoning questions often start with phrases such as "what might be" or "what can be". This is because complex reasoning questions typically place more emphasis on the reasons behind phenomena.

In Table \ref{tab:q_diversity}, we quantify diversity using the average cosine distance between questions in the dataset. Unlike A-OKVQA \cite{schwenk2022okvqa}, we remove punctuation from the questions before using the sentence transformer \cite{jain2022hugging}, and then calculate the average cosine distance among all pairs in the dataset. 



\begin{table}[hbtp]
    \small
    \centering
    \begin{tabular}{   
    l 
    S[table-format=4.1] 
    c
    S[table-format=1.3] 
    S[table-format=1.3]} 
    \toprule
    {Dataset} & {unique instance} & {Avg. length} & {Mean Q distance} \\
    & & {(Q/A)} & & \\ 
    \midrule
    Conversation (LLaVA)  & 256.9k & 10.2/47.4 & 0.845  \\
    Conversation (COCO)  & 36.8k & 8.4/10.4 & 0.721   \\
    Conversation (Objects365)  & 1.7M & 8.5/13.0 & 0.676 \\
    \midrule
    Detail Description (LLaVA)  & 23k & 8.0/104.9 & {-}  \\
    Detail Description (COCO)  & 36.6k & 9.8/157.3 & {-} \\
    Detail Description (Objects365) & 1.7M & 10.0/176.6 & {-}  \\
    \midrule
    Complex Reasoning (LLaVA)  & 76.6k & 12.8/117.8 & 0.845   \\
    Complex Reasoning (COCO)  & 36.8k & 11.9/121.8 & 0.830   \\
    Complex Reasoning (Objects365) & 1.7M & 11.6/116.3 & 0.810 \\
    \bottomrule
    \end{tabular}
    \caption{Statistical analysis of LLaVA data generated by VIGC}
    \label{tab:q_diversity}
\end{table}

\subsection{D. Data Visualization}
\begin{figure}[ht]
\begin{center}
	\includegraphics[width=0.99 \linewidth]{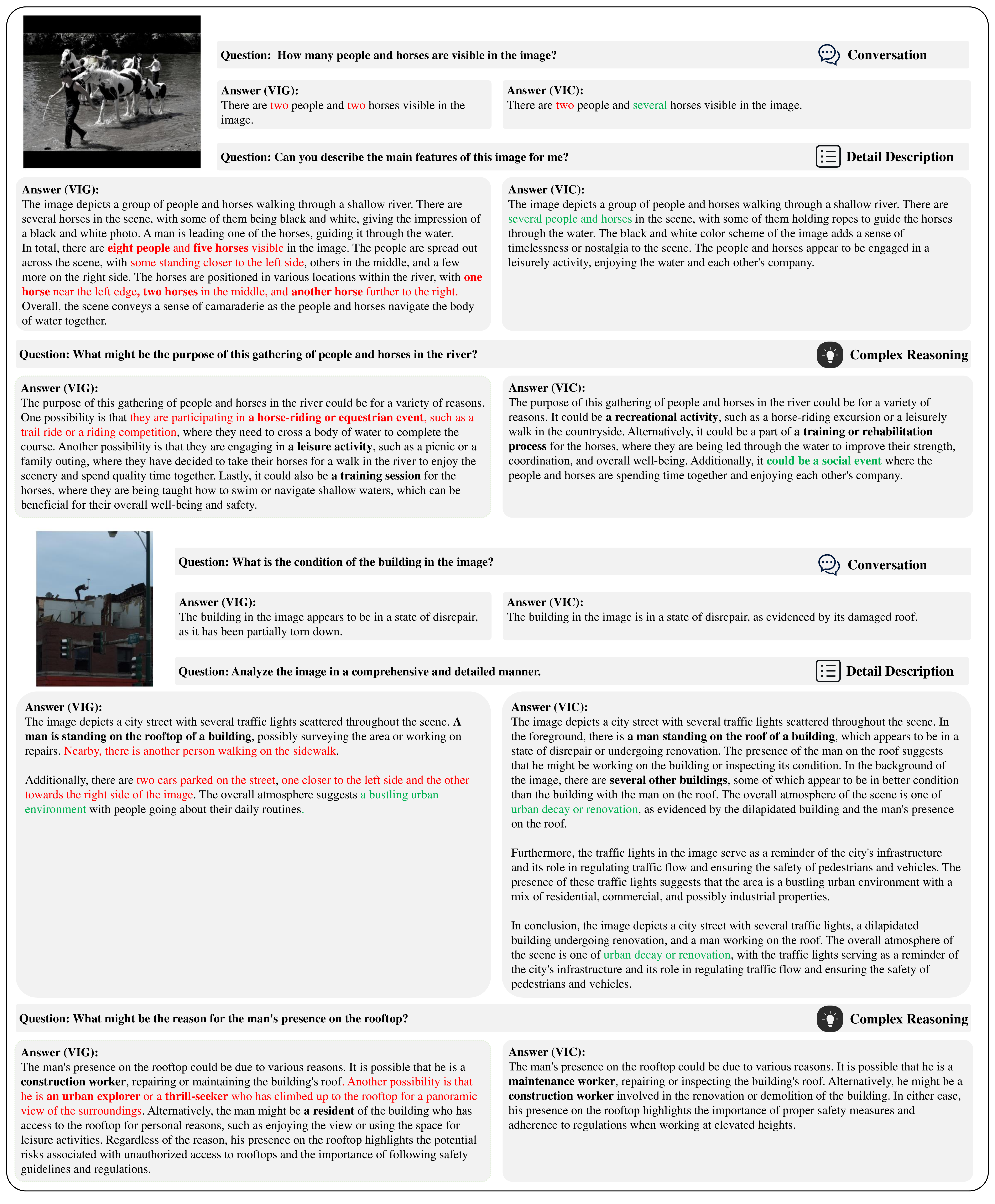}
\end{center}
		\caption{Generated instruction examples based on the proposed VIGC. }
\label{fig:llava_demo2}
\end{figure}

\begin{figure}[ht]
\begin{center}
	\includegraphics[width=0.95 \linewidth]{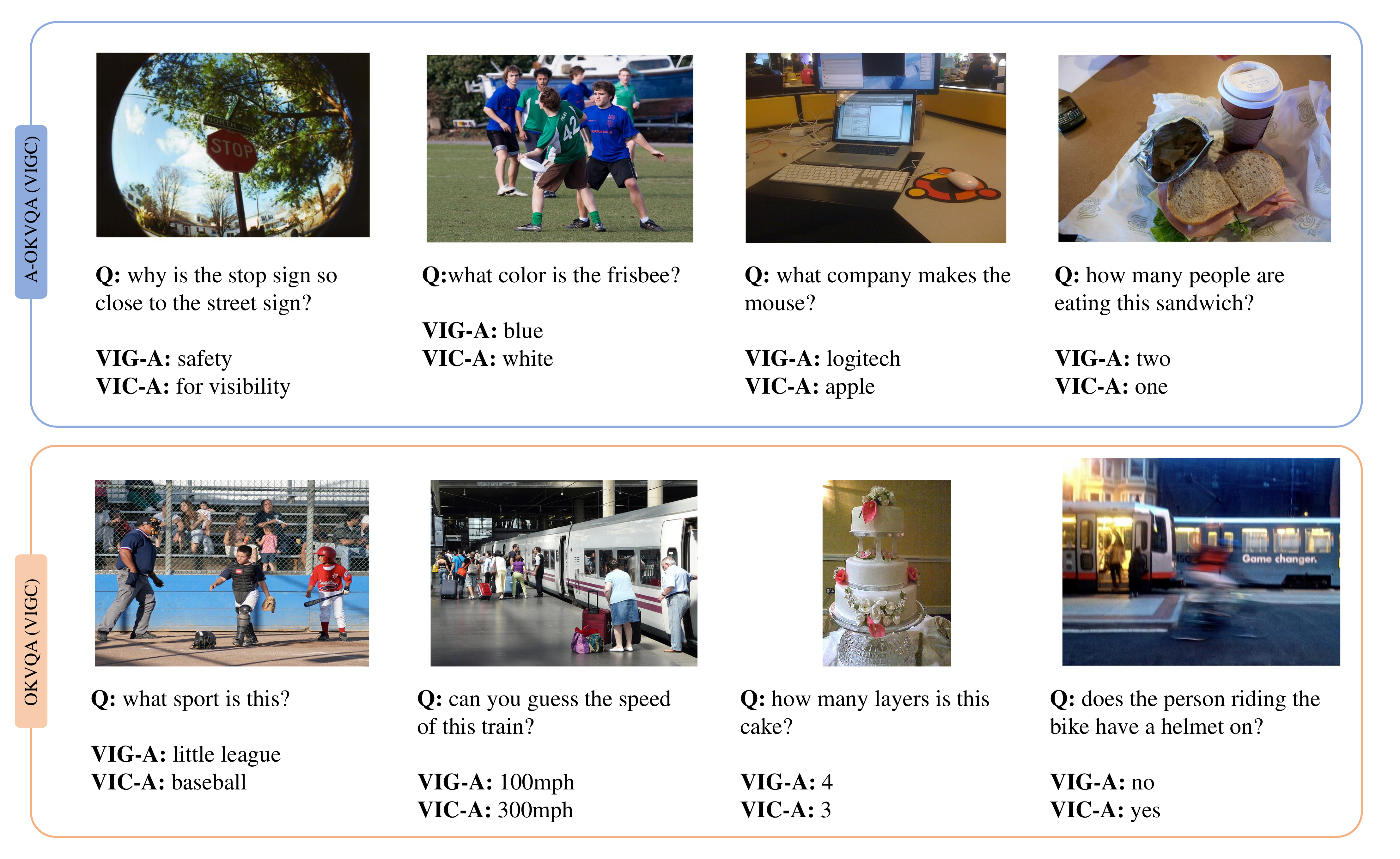}
\end{center}
		\caption{OKVQA and A-OKVQA QA-pairs generated by VIG and VIC}
\label{fig:vqa_vig_vic}
\end{figure}


\begin{figure}[ht]
\begin{center}
	\includegraphics[width=0.95 \linewidth]{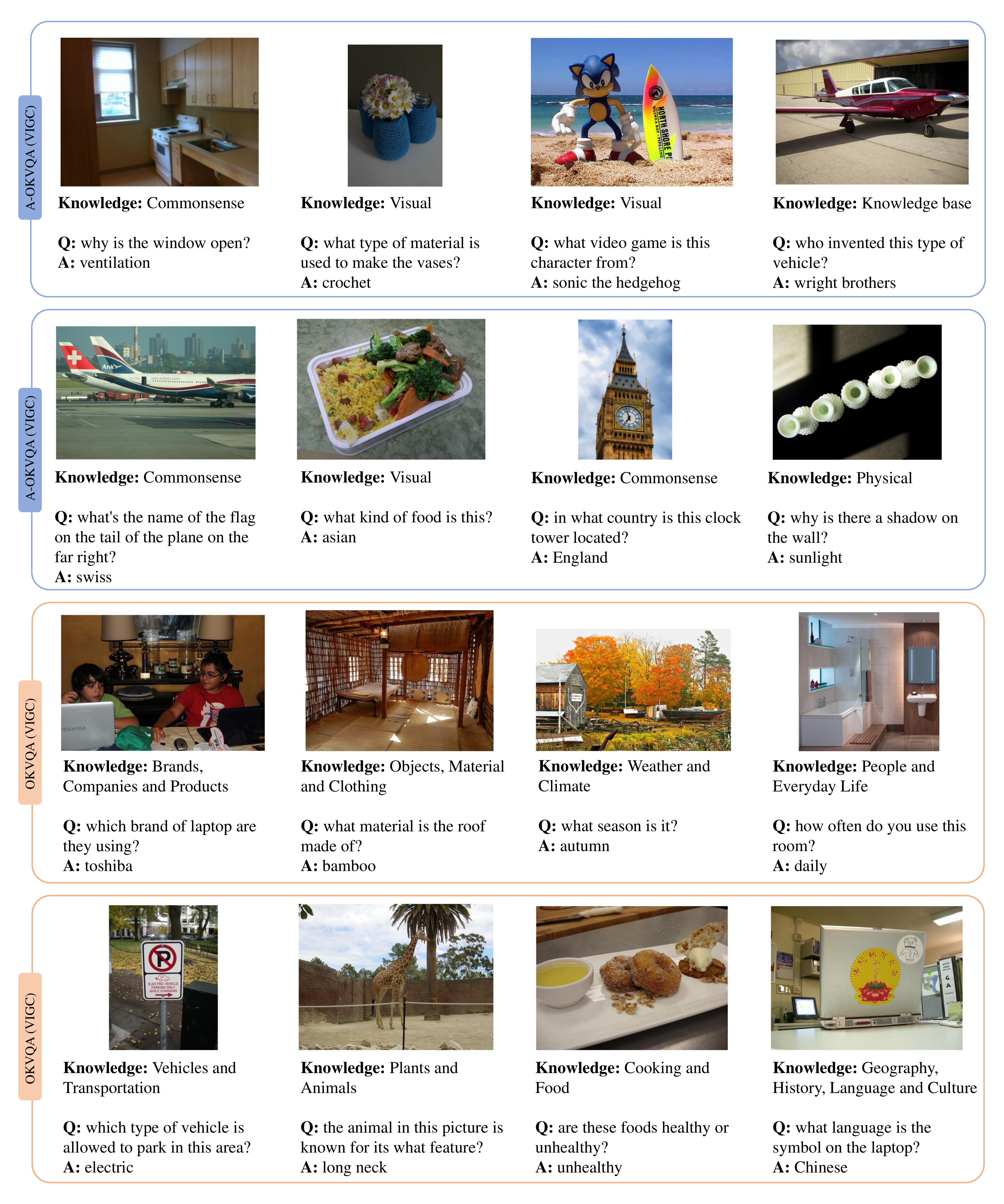}
\end{center}
		\caption{OKVQA and A-OKVQA QA-pairs generated by VIGC}
\label{fig:vqa_demo}
\end{figure}


\end{document}